%% file: dnre_cobb_2023.tex
\documentclass[letterpaper]{article} 
\usepackage{aaai24_arXiv}  
\usepackage{times}  
\usepackage{helvet}  
\usepackage{courier}  
\usepackage[hyphens]{url}  
\usepackage{graphicx} 
\usepackage{natbib}  
\usepackage{caption} 
\usepackage{algorithm}
\usepackage{algorithmic}

\usepackage{newfloat}
\usepackage{listings}
\DeclareCaptionStyle{ruled}{labelfont=normalfont,labelsep=colon,strut=off} 
\floatstyle{ruled}
\newfloat{listing}{tb}{lst}{}
\floatname{listing}{Listing}
\usepackage{booktabs}       
\usepackage{bm}
\usepackage{xcolor} 
\usepackage{graphicx}
\usepackage{caption}
\usepackage{subcaption}
\graphicspath{{./images/}}
\usepackage{multirow}
\usepackage{longtable}
\usepackage{dirtree}
\usepackage{pdfpages}
\usepackage{amsmath}
\usepackage{amssymb}
\usepackage{enumitem}

\title{Direct Amortized Likelihood Ratio Estimation}
\author{
    Adam D. Cobb, Brian Matejek, Daniel Elenius, Anirban Roy, Susmit Jha
}
\affiliations{
    Neuro-Symbolic Computing and Intelligence (NuSCI) Research Group\\
  Computer Science Laboratory, 
  SRI International\\

    adam.cobb@sri.com
}

\begin{document}

\maketitle

\begin{abstract}
  We introduce a new amortized likelihood ratio estimator for likelihood-free simulation-based inference (SBI). Our estimator is simple to train and estimates the likelihood ratio using a single forward pass of the neural estimator. Our approach directly computes the likelihood ratio between two competing parameter sets which is different from the previous approach of comparing two neural network output values. We refer to our model as the direct neural ratio estimator (DNRE). As part of introducing the DNRE, we derive a corresponding Monte Carlo estimate of the posterior. We benchmark our new ratio estimator and compare to previous ratio estimators in the literature. We show that our new ratio estimator often outperforms these previous approaches. As a further contribution, we introduce a new derivative estimator for likelihood ratio estimators that enables us to compare likelihood-free Hamiltonian Monte Carlo (HMC) with random-walk Metropolis-Hastings (MH). We show that HMC is equally competitive, which has not been previously shown. Finally, we include a novel real-world application of SBI by using our neural ratio estimator to design a quadcopter. Code is available at \url{https://github.com/SRI-CSL/dnre}.


\end{abstract}

\input{introduction}

\input{related_work}

\input{method}

\input{experiments}

\input{conclusion}

\section*{Acknowledgements}

This material is based upon work supported by the United States Air Force and DARPA under Contract No. FA8750-20-C-0002.  Any opinions, findings and conclusions or recommendations expressed in this material are those of the author(s) and do not necessarily reflect the views of the United States Air Force and DARPA.

\bibliography{aaai24}

\appendix

\input{appendix}

\end{document}

%% file: introduction.tex
\section{Introduction}

In many scientific applications we rely on complex simulators to provide us with a set of observations, $\mathbf{x}$, for a corresponding set of parameters, $\bm{\theta}$. Common examples range from simulating Computational Fluid Dynamics (or CFD), to flight simulators and computational biology. Significant domain expertise goes into building such simulators to determine the potential functional relationship between $\bm{\theta}$ and $\mathbf{x}$. However, we often do not know the likelihood function, $p(\mathbf{x}|\bm{\theta})$, which is the probability of the output given the input. Without the analytical form of the likelihood, we cannot easily rely on the machinery of Bayesian inference to perform model inversions. For example, given a certain desired output or observation, we might want to know which distribution of parameters could give us this output. In particular, if one were designing a cyber-physical system such as a quadcopter, then a reasonable question to ask could be: ``What size should I make the fuselage such that I achieve a hover time of $20$ seconds?'' If we only have access to a simulator that goes from fuselage parameters to hover time, then this inversion can be challenging. Likelihood-free inference aims to solve this problem of performing inference over simulators (or black-box models). This paradigm is often referred to as simulation-based inference (SBI) \cite{cranmer2020frontier}.

Since one of the main objectives of SBI is to perform Bayesian inference to learn the posterior over the parameters of a simulator, all solutions require accounting for a lack of an analytical likelihood. They revolve around Bayes' rule and the corresponding Bayesian inference approaches that currently exist in the literature. One approach is to directly estimate the posterior using a density estimator, which is referred to as Neural Posterior Estimation \citep{papamakarios2016fast}. NPE learns the posterior distribution using a normalizing flow \citep{rezende2015variational} in a way that allows one to both evaluate and sample from $p(\bm{\theta}|\mathbf{x})$. Unlike NPE, an alternative is Neural Likelihood Estimation (NLE) \citep{papamakarios2019sequential}, which trains a neural density estimator to learn the likelihood directly. The result is the estimated $p(\mathbf{x}|\bm{\theta})$ which can be used with the prior, $p(\bm{\theta})$ to sample from the posterior through Markov chain Monte Carlo (MCMC). Neural Ratio Estimation (NRE) is the third neural-based approach and is the one explored in this paper. Unlike the other two estimation approaches, NRE does not use a density estimator and relies on a standard neural network that aims to estimate the ratio of two likelihoods \citep{cranmer2015approximating, thomas2016likelihood, hermans2020likelihood}. NRE is closely tied to MCMC in that it is used to estimate the ratio in the MH step. While there have been many papers that have focused on comparison between all aforementioned approaches, including their sequential and amortized variants (e.g. \citet{lueckmann2021benchmarking}), the focus of this paper is to explore within the existing range of available ratio estimators.

Our paper is structured as follows. In the next section we highlight related work. In Section \ref{sec:pre} we provide preliminary theory on likelihood ratio estimation. We then introduce DNRE along with likelihood-free HMC in Section \ref{sec:dnre}. Finally we perform an empirical evaluation of the performance of DNRE compared to previous ratio estimators in Section \ref{sec:exp}, leading up to a novel real-world example of designing a quadcopter. We conclude in Section \ref{sec:con}. 

\paragraph{Contributions.} In this paper we build on the work of \citet{hermans2020likelihood} and \citet{thomas2016likelihood, thomas2022likelihood} by introducing a new amortized neural likelihood ratio estimator that \textit{directly} computes the likelihood ratio between two sets of parameters and only requires a single pass through the network to achieve this estimation. We derive a corresponding new Monte Carlo estimate of the posterior distribution when using DNRE. We also introduce a new gradient estimator that can be applied to both our approach and previous approaches. This gradient estimator is more numerically stable than the previous one. We benchmark DNRE along with the baselines both with and without HMC on standard SBI tasks showing that DNRE can often outperform the baselines. These experiments also show likelihood-free HMC to be competitive, which has not been previously shown. Our final contribution is the introduction of a novel design example for quadcopters. 

%% file: related_work.tex
\section{Related Work}

Neural ratio estimation has become one of the main neural network based approaches to performing SBI, alongside NPE and NLE. One of the early works in bringing neural ratio estimation into the area of SBI was by \citet{cranmer2015approximating}. This work showed that likelihood ratios are invariant under a specific class of dimensionality reduction maps. This led to the use of classifiers for approximating the likelihood ratio statistic. The result of this work was to perform likelihood-free inference for ratios between a freely varying parameter and a reference parameter and has since been demonstrated across multiple domains such as in high-energy physics \cite{baldi2016parameterized}. See the corresponding code library, \textit{carl}, for further insights \citep{carl}.

In follow-up work \cite{thomas2016likelihood, thomas2022likelihood, hermans2020likelihood} it was highlighted that relying on a ratio estimation that requires a specific reference parameter, $\bm{\theta}_{\text{ref}}$, can lead to scenarios where the denominator of the ratio, $p(\mathbf{x}|\bm{\theta})/p(\mathbf{x}|\bm{\theta}_{\text{ref}})$, may not provide sufficient coverage for the numerator while exploring $\bm{\theta}$. In particular, in the works by \citeauthor{thomas2016likelihood}, they focus on a logistic regression model and directly show how the approach of requiring a reference parameter can cause large variances in the ratio estimation. \citet{hermans2020likelihood} also highlight this issue with results on their tractable benchmark problem in their paper. Both these works derive a likelihood-to-evidence ratio estimator as their solution to this problem. The idea being that the support of the evidence, $p(\mathbf{x})$, will have better coverage of a likelihood while varying the parameter, $\bm{\theta}$. This likelihood-to-evidence ratio approach is now the one adopted within the SBI community and will be described in Section \ref{sec:pre}.

In the original work by \citet{cranmer2015approximating}, they highlight that it might be possible to learn a classifier which is a function of both $\bm{\theta}$ and $\bm{\theta}_\text{ref}$, as well as $\mathbf{x}$. However, in the description of this approach they write ``it is not clear whether the optimal decision function can be expected for data generated from [two sets of parameters] never jointly encountered during learning'' and therefore do not go with this approach. \citet{hermans2020likelihood} also paraphrase this predicted likely outcome of jointly training pairs of parameters as being ``impractical''. While these works suggest a difficulty in jointly learn pairs, neither of these papers directly explore whether this challenge exists in practice. Our work directly explores this research question and shows that it is feasible to build a ratio estimator parameterized by parameter pairs and often can outperform the likelihood-to-evidence ratio approach.

%% file: method.tex
\section{Preliminaries}\label{sec:pre}
The likelihood ratio between two hypotheses $\bm{\theta}$ and $\bm{\theta}_{\text{ref}}$ for an observation $\mathbf{x}$ is given by:
\begin{equation}\label{eq:TrueRatio}
    r(\mathbf{x}|\bm{\theta}, \bm{\theta}_{\text{ref}}) = \frac{p(\mathbf{x}|\bm{\theta})}{p(\mathbf{x}|\bm{\theta}_{\text{ref}})}.
\end{equation}
\citet{cranmer2015approximating} showed that a change of variables allows one to learn a function, $\mathrm{d}(\mathbf{x})$, to classify between samples, $\mathbf{x}\sim p(\mathbf{x}|\bm{\theta})$, and samples, $\mathbf{x}\sim p(\mathbf{x}|\bm{\theta}_{\text{ref}})$, by providing the classifier with corresponding 1 and 0 labels respectively. We note that this formulation means the classifier is indirectly dependent on $\bm{\theta}$ and $\bm{\theta}_{\text{ref}}$ through the data generation process. The optimal decision function is therefore given by:
\begin{equation}
   \mathrm{d}^*(\mathbf{x}) = \frac{p(\mathbf{x}|\bm{\theta})}{p(\mathbf{x}|\bm{\theta}) + p(\mathbf{x}|\bm{\theta}_{\text{ref}})}.
\end{equation}
The likelihood ratio is then defined using the classifier,
\begin{equation}
   r(\mathbf{x}|\bm{\theta}, \bm{\theta}_{\text{ref}}) = \frac{\mathrm{d}(\mathbf{x})}{1- \mathrm{d}(\mathbf{x})}.
\end{equation}
Instead of the classifier implicitly depending on $\bm{\theta}$, an extension is to directly pass the numerator's parameters to the classifier, while holding $\bm{\theta}_{\text{ref}}$ as a reference parameter \cite{cranmer2015approximating, baldi2016parameterized}. The resulting classifier is now written as $\mathrm{d}(\mathbf{x}, \bm{\theta})$, but is still indirectly dependent on the reference $\bm{\theta}_{\text{ref}}$.

To overcome the challenge of selecting an appropriate reference hypothesis, both \citet{thomas2016likelihood} and \citet{hermans2020likelihood} derive the likelihood-to-evidence ratio,
\begin{equation}\label{eq:orig_ratio}
    r(\mathbf{x}|\bm{\theta}) = \frac{\mathrm{d}^*(\mathbf{x},\bm{\theta})}{1 - \mathrm{d}^*(\mathbf{x},\bm{\theta})} = \frac{p(\mathbf{x},\bm{\theta})}{p(\mathbf{x})p(\bm{\theta})} = \frac{p(\mathbf{x}|\bm{\theta})}{p(\mathbf{x})},
\end{equation}
where $\mathrm{d}^*(\mathbf{x}, \bm{\theta})$ still directly depends on $\bm{\theta}$, but the denominator is now the evidence $p(\mathbf{x})$ and therefore avoids the implicit dependence of the reference parameter. Training this classifier is where \citet{thomas2016likelihood} and \citet{hermans2020likelihood} differ. The former requires sampling from the marginal, whereas the latter only uses samples from the joint. With the latter approach, the learned classifier is trained to distinguish between sample-parameter pairs $(\mathbf{x},\bm{\theta}) \sim p(\mathbf{x},\bm{\theta})$ and independent sample pairs $(\mathbf{x},\bm{\theta}) \sim p(\mathbf{x})p(\bm{\theta})$. As part of this approach \citet{hermans2020likelihood} sample two independent $\bm{\theta}$'s from the prior, $\{\bm{\theta},\bm{\theta}'\}$, and simulate $\mathbf{x}$ according to $p(\mathbf{x}|\bm{\theta})$. This gives the sample from the joint, $(\mathbf{x},\bm{\theta})$, and the independent pair, $(\mathbf{x},\bm{\theta}')$. To avoid the additional cost of drawing a new independent sample $\bm{\theta}'$ the paper applies a shift to the existing vector of $\bm{\theta}$'s to take advantage of the independence between the prior samples. The parameterized classifier is trained using a binary cross entropy loss whereby pairs $(\mathbf{x},\bm{\theta})$ are given the label $1$ and $(\mathbf{x},\bm{\theta}')$ are given the label $0$. The output of the network is an estimate of the log ratio, $\log \hat{r}$. 

Once learned, the likelihood-to-evidence ratio estimator can also be used to estimate the posterior directly by multiplying the ratio by the prior, $p(\bm{\theta}|\mathbf{x}) \approx \hat{r}(\mathbf{x}|\bm{\theta}) p(\bm{\theta})$. However, to estimate the likelihood ratio as in Equation \eqref{eq:TrueRatio}, one needs to apply two forward passes through the network with two sets of parameters to get $r(\mathbf{x}|\bm{\theta}, \bm{\theta}') = r(\mathbf{x}|\bm{\theta})/r(\mathbf{x}|\bm{\theta}')$. This form can then be used to perform likelihood-free MCMC by replacing the likelihood ratio inside the MH acceptance step.

\section{DNRE: Direct Amortized Neural Likelihood Ratio Estimation}\label{sec:dnre}

In this section we present our approach of Direct Amortized Neural Likelihood Ratio Estimation (DNRE), which directly parameterizes the classifier with the two parameter pairs, $\bm{\theta}$ and $\bm{\theta}'$, as $\mathrm{d}(\mathbf{x},\bm{\theta},\bm{\theta}')$. As a result we introduce the new optimal classifier, $\mathrm{d}^*$ as follows:
\begin{align}\label{eq:direct_class}
    \mathrm{d}^*(\mathbf{x},\bm{\theta},\bm{\theta}') =& \frac{p(\mathbf{x},\bm{\theta})p(\bm{\theta}')}{p(\mathbf{x},\bm{\theta})p(\bm{\theta}') + p(\mathbf{x}, \bm{\theta}')p(\bm{\theta})} \notag\\ =& \frac{p(\mathbf{x}|\bm{\theta})p(\bm{\theta})p(\bm{\theta}')}{p(\mathbf{x}|\bm{\theta})p(\bm{\theta})p(\bm{\theta}') + p(\mathbf{x}| \bm{\theta}')p(\bm{\theta}')p(\bm{\theta})}\notag\\ =& \frac{p(\mathbf{x}|\bm{\theta})}{p(\mathbf{x}|\bm{\theta}) + p(\mathbf{x}| \bm{\theta}')},
\end{align}
which leads to the new direct amortized likelihood estimator:
\begin{equation}\label{eq:dir_ratio}
    r(\mathbf{x}|\bm{\theta}, \bm{\theta}') = \frac{\mathrm{d}^*(\mathbf{x},\bm{\theta}, \bm{\theta}')}{1 - \mathrm{d}^*(\mathbf{x},\bm{\theta}, \bm{\theta}')} = \frac{p(\mathbf{x}|\bm{\theta})}{p(\mathbf{x}|\bm{\theta}')}.
\end{equation}

A potential advantage of this approach is the additional information provided to the classifier for label $y=0$. This is when the denominator, $p(\mathbf{x}|\bm{\theta}')$, must be greater than the numerator, $p(\mathbf{x}|\bm{\theta})$. We force this relationship to be the case by swapping $\bm{\theta}$ and $\bm{\theta}'$ such that the parameter in the denominator becomes the one that generated the observation $\mathbf{x}$. This is compared to learning to distinguish between the joint distribution and independent samples which will result in a softer decision boundary for the classifier to learn.

The key difference when training the new Direct Neural Ratio Estimator (DNRE), compared to the original NRE is that during the training, we pass the set of three inputs $(\mathbf{x},\bm{\theta},\bm{\theta}')$, instead of the two. We ensure that the order is consistent while training using the binary cross entropy loss. In our case we assign the ordered triplet $(\mathbf{x},\bm{\theta},\bm{\theta}')$ a label $1$, and swap $\bm{\theta}$ and $\bm{\theta}'$ for the label $0$. We highlight that by explicitly incorporating both sets of parameters we aim to learn the optimal direct likelihood ratio estimator $r(\mathbf{x}|\bm{\theta}, \bm{\theta}')$ in Equation \eqref{eq:dir_ratio}. Algorithm \ref{alg:DNRE} displays the training loop for the new estimator. The algorithm follows that of \cite{hermans2020likelihood}, except for two key differences: (1) including both $\bm{\theta}$ and $\bm{\theta}'$ as estimator inputs and (2) swapping the inputs for the zero label in the binary cross entropy loss.

\begin{algorithm}
\caption{Optimization of $\mathbf{d}_{\bm{\phi}}\label{alg:DNRE}(\mathbf{x},\bm{\theta},\bm{\theta}')$}
\textbf{Inputs}: 
\begin{tabular}{ll}
     $\ell$:& Criterion (BCE) \\
     $p(\mathbf{x}|\bm{\theta})$:& Implicit generative model \\
     $p(\bm{\theta})$:& Prior \\
     $N$:& Number of steps \\
     $M$:& Batch-size
\end{tabular}\\
\textbf{Output}: $\mathbf{d}_{\bm{\phi}}(\mathbf{x},\bm{\theta},\bm{\theta}')$: Parameterized classifier\\
\begin{algorithmic}[1]
\FOR{$i=1$ to $N$}
    \STATE $\bm{\theta} \gets \{\bm{\theta}_m \sim p(\bm{\theta})\}_{m=1}^M$
    \STATE $\bm{\theta'} \gets \{\bm{\theta'}_m \sim p(\bm{\theta})\}_{m=1}^M$
    \STATE $\mathbf{x} \gets \{\mathbf{x}_m \sim p(\mathbf{x}|\bm{\theta})\}_{m=1}^M$
    \STATE $\mathcal{L} \gets \ell (\mathbf{d}_{\bm{\phi}}(\mathbf{x},\bm{\theta},\bm{\theta}'), 1) + \ell ( \mathbf{d}_{\bm{\phi}}(\mathbf{x},\bm{\theta}',\bm{\theta}), 0)$ 
    \STATE $\bm{\phi} \gets \text{OPTIMIZER}(\bm{\phi}, \nabla_{\bm{\phi}}\mathcal{L})$
\ENDFOR
\STATE \textbf{return} $\mathbf{d}_{\bm{\phi}}$
\end{algorithmic}
\end{algorithm}

\subsection{Monte Carlo Posterior Approximation}

Unlike the likelihood-to-evidence ratio proposed in Equation \eqref{eq:orig_ratio}, the DNRE only requires one pass through the neural classifier to estimate the likelihood ratio. This means when performing MCMC, one only needs to call the classifier once per evaluation of the MH acceptance step. This direct estimation is therefore more convenient. However, a significant advantage that the original estimator has over our proposed estimator, is that the posterior distribution can be explicitly approximated by adding the log prior to the predicted log likelihood-to-evidence ratio. 

To overcome the challenge of estimating the posterior using DNRE, we must integrate out the $\bm{\theta}'$ in the denominator numerically, using $M$ samples. This requires an inverse trick that we derive in the log-space:
\begin{align}\label{eq:dnre_post}
    \log p(\bm{\theta}|\mathbf{x}) \approx - \mathrm{logSumExp}\bigl\{-\log \hat{r}(\mathbf{x}|\bm{\theta}, \bm{\theta}_i')) \bigr\} \notag\\  + \log M + \log p(\bm{\theta}).
\end{align}
The $\mathrm{logSumExp}$ is applied to the negated output, $\log \hat{r}$, of the neural network to estimate the log of the inverse of the likelihood-to-evidence ratio ($p(\mathbf{x})/p(\mathbf{x}|\bm{\theta})$). The summation adds a $\log M$ term, which is removed in the log-space. A final negation reverses the initial negation, and gives the estimated likelihood-to-evidence ratio. Then, adding the log-prior to gives the log posterior. We show this derivation in full starting from the exponentiated output of the predicted DNRE, $r(\mathbf{x}|\bm{\theta}, \bm{\theta}_i')$:
\begin{align}
    \frac{1}{p(\bm{\theta}|\mathbf{x})} &\approx \frac{1}{p(\bm{\theta})}\frac{1}{M}\sum_i^M\frac{1}{r(\mathbf{x}|\bm{\theta}, \bm{\theta}'_i)} \notag\\
    &= \frac{1}{p(\bm{\theta})}\frac{1}{M}\sum_i^M\frac{p(\mathbf{x}|\bm{\theta}'_i)}{p(\mathbf{x}|\bm{\theta})} \notag\\
    &=\frac{1}{p(\bm{\theta})}\frac{p(\mathbf{x})}{p(\mathbf{x}|\bm{\theta})}. \notag
\end{align}

For higher-dimensional $\bm{\theta}'$ this is computationally intractable, but for smaller dimensional problems we can use this form to compare posteriors in a manner consistent with previous approaches within the literature. In particular, this form allows us to analyze how the expected coverage \cite{hermans2021averting} varies with the sample size $M$. We will explore this in the experiments section.

\subsection{Likelihood-free Hamiltonian Monte Carlo}

Although not empirically explored in \citet{hermans2020likelihood}, they make a recommendation as to how one could implement Hamiltonian Monte Carlo (HMC) using a neural ratio estimator. HMC is a gradient-based Markov chain Monte Carlo sampling scheme that augments the original parameter space with additional momentum parameters, $\mathbf{m}$, in order to sample using Hamiltonian dynamics \citep{duane1987hybrid, neal2011mcmc}. HMC defines the potential energy function as $U(\bm{\theta}) = - \log[p(\mathbf{x}|\bm{\theta})p(\bm{\theta})]$ and the kinetic energy function as $K(\mathbf{m}) = \mathbf{m}^{\top}\mathbf{m}/2 $. The two requirements of running an HMC algorithms are access to $\nabla U(\bm{\theta})$ and access to the MH acceptance step of $\rho = \min(0, -U(\bm{\theta}^*) + U(\bm{\theta}) -K(\mathbf{m}^*) + K(\mathbf{m}) ) $, where $\bm{\theta}^*$ and $\mathbf{m}^*$ are the proposed parameter and momenta pair. For NRE and BNRE, $\rho$ is derived from two passes through the estimator such that $-U(\bm{\theta}^*) + U(\bm{\theta}) = \log r(\mathbf{x}|\bm{\theta}^*) -  \log r(\mathbf{x}|\bm{\theta}) + \log p(\bm{\theta}^*) - \log p(\bm{\theta})$. For our new estimator DNRE, we only require one pass such that $-U(\bm{\theta}^*) + U(\bm{\theta}) = \log r(\mathbf{x}|\bm{\theta}^*, \bm{\theta}) +\log p(\bm{\theta}^*) -\log p(\bm{\theta})$. To estimate $\nabla U(\bm{\theta})$, \citet{hermans2020likelihood} use the chain rule to derive:
\begin{equation}\label{eq:old_deriv}
    \nabla_{\bm{\theta}} U(\bm{\theta}) = - \frac{\nabla_{\bm{\theta}} r(\mathbf{x}|\bm{\theta})}{r(\mathbf{x}|\bm{\theta})}.
\end{equation}
This estimate of the derivative can be numerically unstable in practice as it requires exponentiating the output of the estimator for the denominator, which we have observed can often be a small number. A more simple approach is to estimate $\nabla U(\bm{\theta})$ by treating the classifer as the approximate log ratio which allows us to separate the two terms. Therefore in the case of the NRE we can achieve the approximated derivative of the log likelihood as:
\begin{align}\label{eq:new_deriv_nre}
    \nabla_{\bm{\theta}} \log r(\mathbf{x}|\bm{\theta}) &\approx \nabla_{\bm{\theta}} \log p(\mathbf{x}|\bm{\theta}) - \nabla_{\bm{\theta}} \log p(\mathbf{x})\notag\\ &= \nabla_{\bm{\theta}} \log p(\mathbf{x}|\bm{\theta}).
\end{align}
We can also follow the same approximation for our DNRE approach to get the derivative of the log likelihood:
\begin{align}\label{eq:new_deriv_dnre}
    \nabla_{\bm{\theta}} \log r(\mathbf{x}|\bm{\theta}, \bm{\theta}') &\approx \nabla_{\bm{\theta}} \log p(\mathbf{x}|\bm{\theta}) - \nabla_{\bm{\theta}} \log p(\mathbf{x}|\bm{\theta}') \notag\\ &=  \nabla_{\bm{\theta}} \log p(\mathbf{x}|\bm{\theta}).
\end{align}
We have found these estimators to be more stable as it avoids the exponential and division operations. For DNRE, we need to account for the $\bm{\theta}'$ when estimating the derivative. Therefore we simply choose to sample $\bm{\theta}'$ from its prior during our implementation. For estimators that closely match the true ratio, the contribution from $\bm{\theta}'$ to the derivative should tend to zero, following the logic of Equation \eqref{eq:new_deriv_dnre}.

%% file: experiments.tex
\section{Experiments}\label{sec:exp}

In this section we compare three amortized estimators: Neural Ratio Estimator (NRE) \cite{hermans2020likelihood}, Balanced Neural Ratio Estimator (BNRE) \cite{delaunoy2022towards}, and our new approach of Direct Neural Ratio Estimator (DNRE). The baseline approaches are implemented using the \textit{LAMPE} Python package \cite{lampe}. BNRE uses the same loss function as NRE, but with an additional regularization term to encourage a more balanced and conservative overall estimation of the posterior. We start with a toy example to assess the ability of modeling a simple ground truth likelihood ratio. We then explore how the Monte Carlo posterior approximation performs in practice using the two moons dataset example. Finally, we compare the performance of all three ratio estimators using a common SBI benchmark. In our final experiment we demonstrate a full scale example for for the design of a quadcopter.

\subsection{Illustrative Ratio Approximation Example}

\begin{figure}[h!]
    \centering
    \includegraphics[width=\columnwidth]{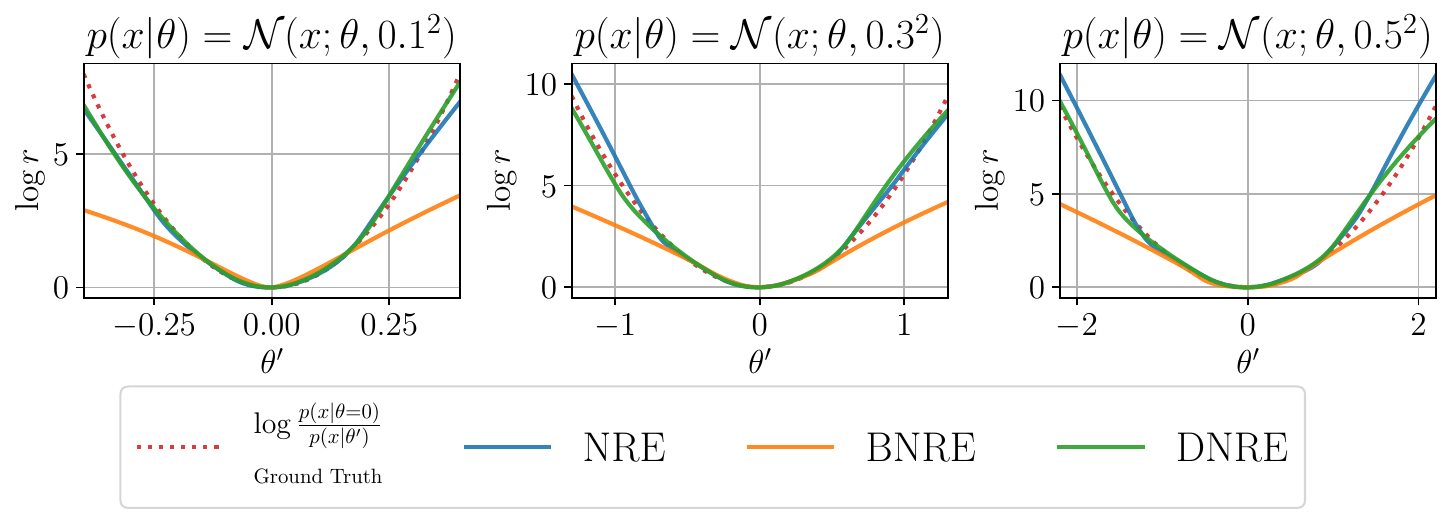}
    \caption{Comparison to ground truth log likelihood ratio for the Gaussian model while varying the denominator parameter $\theta'$ and keeping the numerator parameter constant at $\theta=0.0$. DNRE outperforms NRE, followed by BNRE.}
    \label{fig:gauss}
\end{figure}

Our initial comparison investigates the ability of all three approaches to fit to a simple one-dimensional hierarchical Gaussian model,
\begin{equation}
    x \sim \mathcal{N}(\theta, \sigma^2), \quad \theta \sim \mathcal{N}(0, \sigma^2). \notag
\end{equation}
All models were trained over 1000 epochs with neural network architectures of three layers of 64 units. We use a training set of size $10{,}000$ and a validation set of size $5{,}000$. For comparison to the ground truth, we set the numerator to $p(x|\theta=0.0)$ and set the $\theta'$ in the denominator to range from the minimum to the maximum of the training set. Figure \ref{fig:gauss} displays the results for three different standard deviations, $[0.1, 0.3, 0.5]$. We see that the DNRE slightly outperforms NRE, which is supported by the results of Table \ref{tab:gauss_res} in Appendix \ref{app:ill}, and BNRE seems to struggle when in regions where the difference between the two log likelihoods is large (greater than around two, which corresponds to a probability ratio of just over seven).  

\subsection{Posterior Approximation Performance}
\begin{figure}[h!]
    \centering
    \includegraphics[width=\columnwidth]{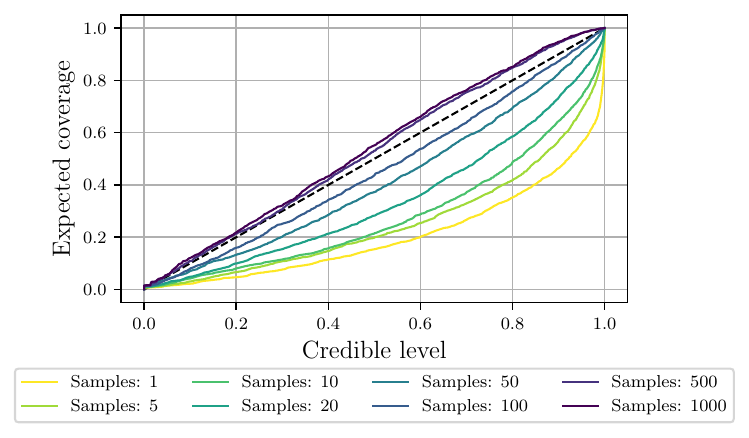}
    \caption{Expected coverage with varying sample size $M$ for the two moons dataset. The larger the sample size, the more balanced the posterior estimate becomes.}
    \label{fig:dnre_coverage}
\end{figure}

For the next experiment, we work with the two moons dataset, which is a typical low-dimensional benchmark within the SBI literature \cite{greenberg2019automatic,lueckmann2021benchmarking}. We explore the quality of the estimator by analyzing our new posterior approximation from Equation \eqref{eq:dnre_post}. Figure \ref{fig:dnre_coverage} displays the expected coverage plot, whereby perfectly calibrated posteriors follow the dashed line, with more conservative posteriors falling above this line and overconfident ones falling below \cite{hermans2021averting}. Conservative estimators are generally more desirable for scientific applications. This figure highlights the importance of the number of samples needed to realize the estimated calibration of the DNRE approximated posterior. Increasing the number of these samples leads to the best performing (slightly conservative) expected coverage performance. 

\begin{figure}[h!]
    \centering
    \includegraphics[width=\columnwidth]{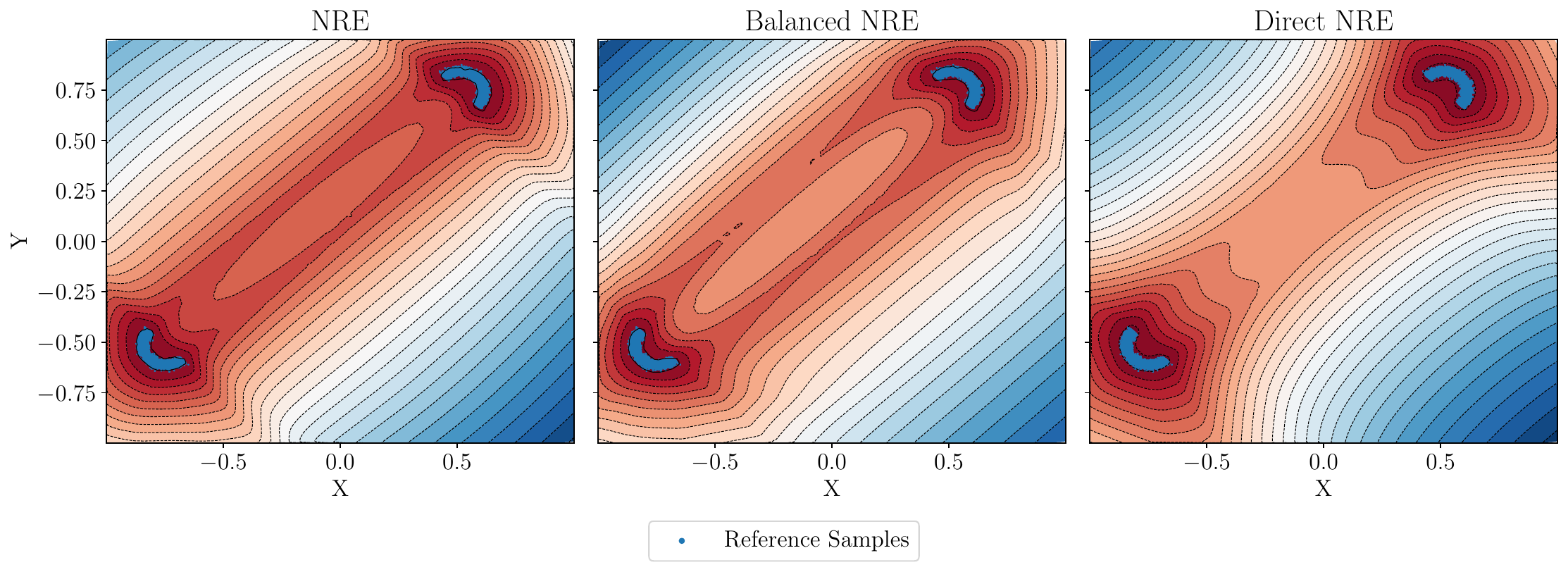}
    \caption{Contour plot comparing the three amortized estimators for the two moons dataset: Neural Ratio Estimator (NRE), Balanced Neural Ratio Estimator (BNRE), and Direct Neural Ratio Estimator (DNRE). For the DNRE, we used $M=20,000$. }
    \label{fig:dnre_post}
\end{figure}

In Figure \ref{fig:dnre_post} we compare the inferred posteriors of NRE, BNRE, and DNRE. We also overlay the reference samples from the ground truth conditional distribution. The main illustration of this figure is to show that our new approach is correctly approximating the posterior distribution. 

\subsection{Benchmark results}\label{sec:bench}

In this section we perform a quantitative comparison between the three ratio estimators, where we use the SBI Benchmark examples of \citet{lueckmann2021benchmarking} (see Appendix \ref{app:bench}). All estimators have an architecture of five layers of $64$ units, using the Exponential Linear Unit non-linearity between layers. For all approaches we applied the same grid search over both the learning rate and standard deviation of the proposal distribution for the random-walk MH sampling scheme. The best model for each task was selected according to the \texttt{C2ST} score applied for $1{,}000$ samples using the first observation of the benchmark. The \texttt{C2ST} score is a classifier-based test trained to distinguish between reference posterior samples and samples from the likelihood-free inference (see Appendix \ref{app:bench}). The results in Table \ref{tab:sbi_results_mh} are averaged across the 10 available observations in the benchmark. A score of $0.5$ is optimal, whereas $1.0$ would be the worst score. While the \texttt{C2ST} metric cannot capture all the key indicators of a good estimator, these results do indicate the promise of DNRE since it outperforms the baselines. We see that DNRE achieves the best score in five out of the ten benchmarks, followed by NRE with four and BNRE with one. 

\begin{table*}
\begin{center}
\begin{scriptsize}
\begin{sc}
\begin{tabular}{lccccc}
\toprule
Approach & TM & GL & LV & SIR & SLCP \\
\midrule
NRE & $0.559 \pm 0.026$ & $\mathbf{0.533 \pm 0.015}$ & $\mathbf{0.995 \pm 0.004}$ & $\mathbf{0.803 \pm 0.083}$ & $0.925 \pm 0.039$\\
BNRE & $\mathbf{0.544 \pm 0.033}$ & $0.59 \pm 0.019$ & $0.997 \pm 0.002$ & $0.944 \pm 0.029$ & $0.893 \pm 0.038$\\
DNRE & $0.587 \pm 0.027$ & $0.576 \pm 0.034$ & $0.998 \pm 0.002$ & $0.871 \pm 0.064$ & $\mathbf{0.826 \pm 0.078}$\\
\toprule
& GLU & SLCP D & B GLM & GM & B GLM R \\
\midrule
NRE & $0.618 \pm 0.025$ & $0.982 \pm 0.008$ & $\mathbf{0.781 \pm 0.046}$ & $0.751 \pm 0.015$ & $0.819 \pm 0.035$\\
BNRE & $0.607 \pm 0.021$ & $0.984 \pm 0.007$ & $0.807 \pm 0.039$ & $0.755 \pm 0.016$ & $0.862 \pm 0.047$\\
DNRE & $\mathbf{0.597 \pm 0.025}$ & $\mathbf{0.98 \pm 0.011}$ & $0.813 \pm 0.069$ & $\mathbf{0.747 \pm 0.015}$ & $\mathbf{0.777 \pm 0.062}$\\
\bottomrule
\end{tabular}
\end{sc}
\end{scriptsize}
\end{center}
\caption{C2ST SBI Metropolis-Hastings Benchmark Results.}
\label{tab:sbi_results_mh}
\end{table*}

\subsection{Comparison to HMC}

In addition to comparing the results for all ratio estimators across the benchmark examples using random-walk MH, we also run HMC using the gradient estimators shown in Equations \eqref{eq:new_deriv_nre} and \eqref{eq:new_deriv_dnre}. We display these results in Table \ref{tab:sbi_results_hmc}. As with the random-walk MH sampler, we perform grid search using the same setup, except we now additionally vary the trajectory length instead of the variance of the proposal distribution. Since HMC can be notoriously difficult to tune, we resort to extending the grid search to include desired acceptance rates ranging between $0.5$ to $0.8$. This uses the dual averaging scheme as introduced in \citet{hoffman2014no} and as implemented in \citet{cobb2019introducing}.\footnote{We additionally average across all chains at each step, which results in reliable desired acceptance rates.} We use the same number of parallel chains; same number of collected samples; and the same thinning of each chain as for the random-walk MH sampler. Again, we see that DNRE achieves the lowest score in five of the ten experiments, however these experiments are interestingly not for the same experiments as in Table \ref{tab:sbi_results_mh}.

\begin{table*}
\begin{center}
\begin{scriptsize}
\begin{sc}
\begin{tabular}{lccccc}
\toprule
Approach & TM & GL & LV & SIR & SLCP \\
\midrule
NRE & $0.657 \pm 0.026$ & $0.531 \pm 0.015$ & $\mathbf{0.996 \pm 0.005}$ & $\mathbf{0.803 \pm 0.082}$ & $0.913 \pm 0.042$\\
BNRE & $\mathbf{0.562 \pm 0.07}$ & $0.543 \pm 0.019$ & $0.997 \pm 0.003$ & $0.945 \pm 0.029$ & $\mathbf{0.891 \pm 0.039}$\\
DNRE & $0.753 \pm 0.14$ & $\mathbf{0.519 \pm 0.006}$ & $0.998 \pm 0.001$ & $0.891 \pm 0.043$ & $0.892 \pm 0.098$\\
\toprule
& GLU & SLCP D & B GLM & GM & B GLM R \\
\midrule
NRE & $0.613 \pm 0.025$ & $0.995 \pm 0.003$ & $0.776 \pm 0.044$ & $0.752 \pm 0.016$ & $0.83 \pm 0.044$ \\
BNRE & $\mathbf{0.598 \pm 0.025}$ & $0.991 \pm 0.009$ & $0.788 \pm 0.036$ & $0.751 \pm 0.015$ & $0.86 \pm 0.047$ \\
DNRE & $0.665 \pm 0.162$ & $\mathbf{0.989 \pm 0.008}$ & $\mathbf{0.738 \pm 0.038}$ & $\mathbf{0.747 \pm 0.013}$ & $\mathbf{0.77 \pm 0.109}$ \\
\bottomrule
\end{tabular}
\end{sc}
\end{scriptsize}
\end{center}
\caption{C2ST SBI HMC Benchmark Results.}
\label{tab:sbi_results_hmc}
\end{table*}

We can also compare inference through HMC with random-walk MH, since it is useful to establish whether either sampling approach is in general more useful on this benchmark. Table \ref{tab:sbi_results_diff} displays the mean \texttt{C2ST} of the MH results minus the mean \texttt{C2ST} of the HMC results. When HMC has a lower (better) \texttt{C2ST}, we highlight this in \textcolor{orange}{orange}, and we do the same for MH by highlighting when it is better using \textcolor{blue}{blue}. A potentially useful result of this table is that there are clear winners for certain tasks. For the two moons and the SLCP Distractors task, random-walk MH outperforms across all estimators, whereas for Gaussian linear and Bernoulli GLM, HMC outperforms across all estimators. This observation could imply that certain SBI tasks are suited to a particular sampling scheme. Additionally, there are thirteen instances of each sampling scheme performing best, further suggesting that both HMC and MH samplers should be considered for all future SBI approaches since we cannot conclude from these results that either sampling scheme is better.

\begin{table}[h!]
\begin{center}
\begin{scriptsize}
\begin{sc}
\begin{tabular}{lccccc}
\toprule
Approach & TM & GL & LV & SIR & SLCP \\
\midrule
NRE & $\textcolor{blue}{-0.098}$ &  $\textcolor{orange}{0.002}$ & $\textcolor{blue}{-0.001}$&  $0.000$   & $\textcolor{orange}{0.012}$\\
BNRE & $\textcolor{blue}{-0.018}$&  $\textcolor{orange}{0.047}$&  $0.000$   & $\textcolor{blue}{-0.001}$&  $\textcolor{orange}{0.002}$\\
DNRE & $\textcolor{blue}{-0.166}$&  $\textcolor{orange}{0.057}$&  $0.000$   & $\textcolor{blue}{-0.020}$ & $\textcolor{blue}{-0.066}$\\
\toprule
& GLU & SLCP D & B GLM & GM & B GLM R \\
\midrule
NRE & $\textcolor{orange}{0.005}$ & $\textcolor{blue}{-0.013}$ &  $\textcolor{orange}{0.005}$ & $\textcolor{blue}{-0.001}$ & $\textcolor{blue}{-0.011}$\\
BNRE &  $\textcolor{orange}{0.009}$& $\textcolor{blue}{-0.007}$&  $\textcolor{orange}{0.019}$&  $\textcolor{orange}{0.004}$&  $\textcolor{orange}{0.002}$\\
DNRE & $\textcolor{blue}{-0.068}$& $\textcolor{blue}{-0.009}$&  $\textcolor{orange}{0.075}$&  $0.000$   &  $\textcolor{orange}{0.007}$\\
\bottomrule
\end{tabular}
\end{sc}
\end{scriptsize}
\end{center}
\caption{C2ST SBI comparison between Table \ref{tab:sbi_results_mh} and Table \ref{tab:sbi_results_hmc}. The results show the means of the MH minus the HMC means. Better performance by \textcolor{orange}{HMC} is given by \textcolor{orange}{orange}. Better performance by \textcolor{blue}{MH} is given by \textcolor{blue}{blue}.}
\label{tab:sbi_results_diff}
\end{table}

\subsection{Estimated Posterior Under Ground Truth $\bm{\theta}$}
The benchmark provided by \citet{lueckmann2021benchmarking} also provides the ground truth, $\bm{\theta}^*$, for each observation. As a result it is possible to report the posterior probability of $\bm{\theta}^*$ according to each estimator. Table \ref{tab:sbi_log_post} in Appendix \ref{app:post} displays the average log posterior probabilities across all observations for each estimator. We see that DNRE tends to assign the highest posterior probability when averaging over all the ground truth observation and parameter pairs.

\subsection{Quadcopter Design}\label{sec:uav}

\begin{figure}[h!]
    \centering
    \includegraphics[width=\columnwidth]{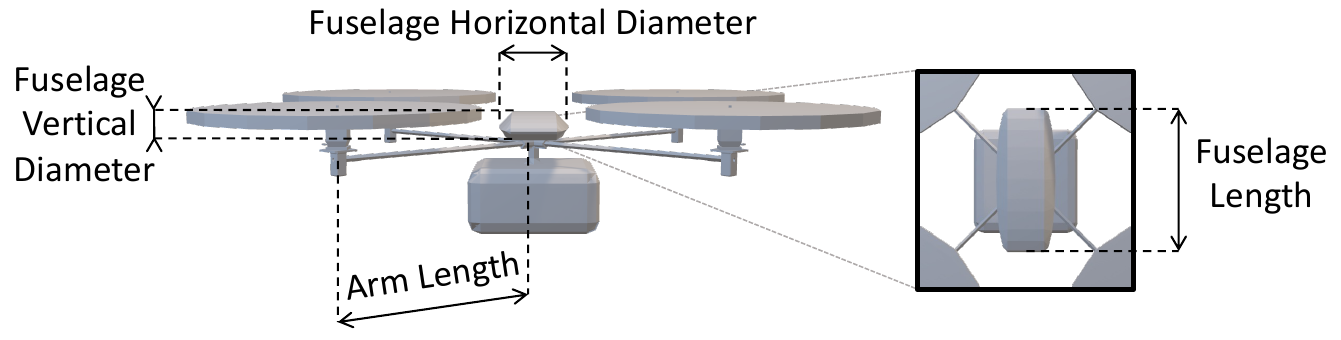}
    \caption{Schematic of the quadcopter design which highlights the key design parameters related to the fuselage and the arm length.}
    \label{fig:uav}
\end{figure}

For our final experiment we perform SBI for aircraft design. In particular, we run a custom flight dynamics pipeline which combines CAD software~\cite{creo} with a computational flight simulation model~\cite{walker2022flight, Bapty2022design, cobb2023aircraft}. To use the pipeline, one can define an aircraft design and allow the simulator to determine performances such as its drag, lift, maximum flight time and hover time. We want to explore how we should parameterize a quadcopter such that we achieve a certain performance level. The observations, $\mathbf{x} \in \mathbb{R}^7$, consist of: the number of interferences; the mass; the maximum flight distance; the maximum hover time; the maximum lateral speed; the maximum control input at the maximum flight distance; and the maximum power at the maximum speed. The design parameters, $\bm{\theta} \in \mathbb{R}^{19}$, consist of: the arm length; four fuselage shape parameters; and seven `x' and `y' locations of electrical devices inside the fuselage. We highlight some of the key parameters in Figure~\ref{fig:uav}.
\subsubsection{Training}
We train all three neural ratio estimators using a five layer fully connected architecture with $128$ hidden units per layer. We use a learning rate of $0.001$ and train all models for $2{,}000$ epochs using a batch size of $512$ with $3{,}592$ training points and $898$ validation points.

\subsubsection{Design Specification}
\begin{figure}[h!]
    \centering
    \includegraphics[width=0.75\columnwidth]{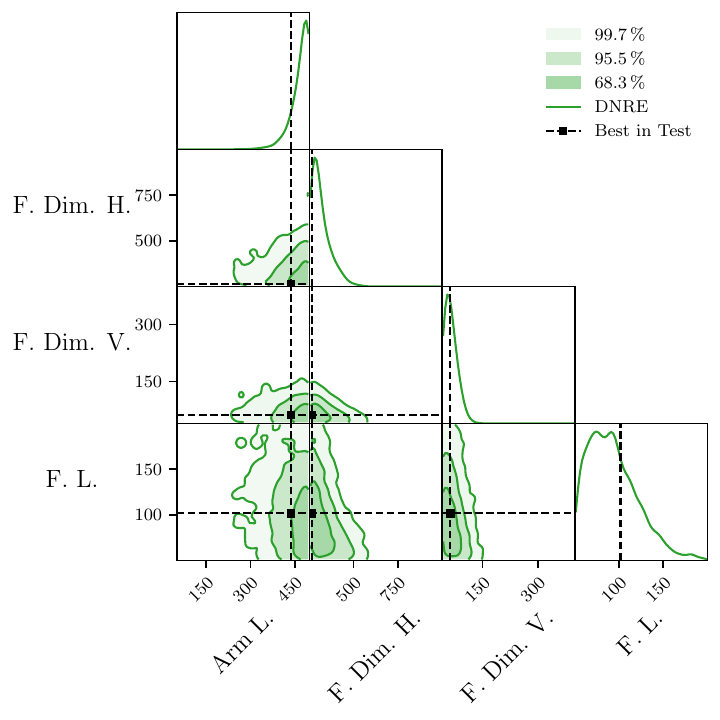}
    \caption{Corner plot displaying materialized samples from likelihood-free inference using DNRE for a subset of the parameters as highlighted in Figure \ref{fig:uav}. The black square corresponds to the highest ranked test design in the test set. The high density on the longest arm length combined with parameters leading to a small fuselage likely follow the objective of reducing interferences between the propellers and reducing the overall mass.}
    \label{fig:uav_dnre}
\end{figure}

SBI is useful tool for design when using black-box models to simulate performance. 
In this case the sole objective is to perform Bayesian inference over the simulation model, such that we can specify objectives ($= \mathbf{x}$) to sample designs and evaluate their suitability. Here we specify a design objective of:
\begin{center}
\begin{scriptsize}
\begin{tabular}{lc|lc}
    Interferences &  $0$ & Max. Flight Distance (m) & $500.0$\\
    Mass (Kg)& $3$ & Max. Hover Time (s) & $20.0$ \\
    Max. Speed (m/s)& $30.0$ & Power at Max. Speed (W) & 2000.0 \\
    Max. Control Input & 0.7 \\
\end{tabular}
\end{scriptsize}
\end{center}
To see how this design objective compares to existing designs in the training data, in the supplementary materials we include Figure~\ref{fig:uav_data} which displays the empirical distribution of the observations along with this specific design objective superimposed. Notably, this is a particularly demanding specification when looking at the need of a low mass quadcopter with a high hover time. Figure \ref{fig:uav_dnre} displays the result of performing likelihood-free inference using random walk MH with the DNRE approach. We also highlight the highest ranked design, according to DNRE, from a randomly sampled test set of $1{,}059$ quadcopters. Interestingly, we see that designs meeting our specification come from a distribution that require long arm lengths and a minimal sized fuselage. We hypothesize that our zero interference requirement has led longer arm lengths to prevent the propellers intersecting with other components. The choice of a small fuselage (short diameters and short length) will help reduce the mass since the average mass of $5.38$~Kg is much higher than our desired design choice. Appendix \ref{app:uav} includes the comparable corner plots for NRE and BNRE.

\subsubsection{Ranking}
We can use the estimated likelihood ratio to perform a model comparison for different $\bm{\theta}$'s. In particular we can use all three ratio estimators to compare all the designs in the test set. We look to find the top $100$ designs according to each likelihood estimator and check for consistency across estimators. The following ranking matrix displays the number of overlapping test samples in both the top $100$ (highlighted in \textcolor{blue}{blue}) and the top $10$ (highlighted in \textcolor{orange}{orange}) out of $1{,}059$ data points:
\begin{center}
\begin{small}
\begin{tabular}{lccc}
    \multicolumn{4}{c}{Overlap: \textcolor{blue}{Top-100}, \textcolor{orange}{Top-10}}\\
    \midrule
        & NRE & BNRE & DNRE \\
    NRE &  -  &  \textcolor{blue}{93}  &  \textcolor{blue}{89}  \\
    BNRE&  \textcolor{orange}{9}  &   -   &  \textcolor{blue}{87}   \\
    DNRE&   \textcolor{orange}{8}  &  \textcolor{orange}{8}    &    -  \\
\end{tabular}
\end{small}
\end{center}
All estimators have a high overlap or design rankings for both top 10 and top 100 suggesting that all approaches are valid. Furthermore, there is a clear `winner' in the test set since all propose the same top performing design which is the one superimposed on Figure \ref{fig:uav_dnre}.

\subsubsection{Improving an Existing Design}
\begin{figure}[h!]
    \centering
    \includegraphics[width=\columnwidth]{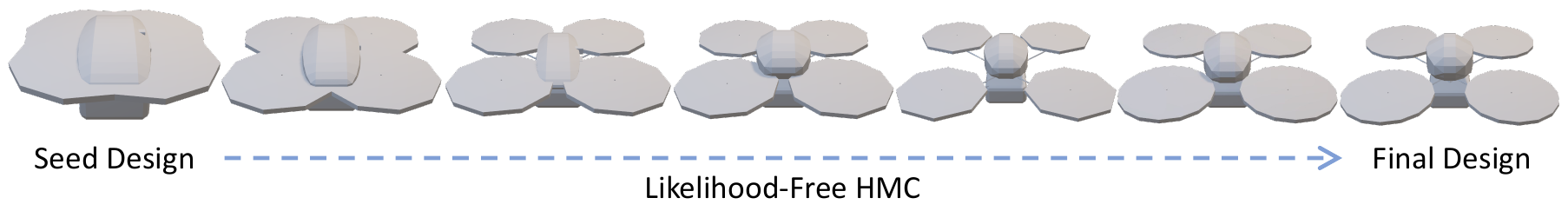}
    \caption{Sub-sampled quadcopter designs taken along the likelihood-free HMC chain using DNRE. The initial seed design on the far left has multiple structural interferences, including sensors that cut through the fuselage. As we move along the chain we see the design morph into our desired structure with very few interferences. This is achieved through increasing the arm length and changing the shape of the fuselage, as well as varying the placement of the interior sensing components.}
    \label{fig:uav_stls}
\end{figure}

We demonstrate the utility of performing likelihood-free HMC with DNRE by taking a promising existing design and improving it by reducing structural interferences. In Appendix \ref{app:uav} we include the full case study that demonstrates how we can use our newly derived gradient-based sampling scheme with our proposed direct neural ratio estimator to evolve a seed design from $60$ structural interferences to $4$. This is while successfully preserving the other original design performance metrics as evaluated through the flight simulation software. Figure \ref{fig:uav_stls} displays sub-sampled quadcopters taken along the Markov chain.

%% file: conclusion.tex
\section{Conclusion}\label{sec:con}
The contribution of this paper is to demonstrate that directly learning to approximate the likelihood ratio between two pairs of parameters presents itself as a viable option for likelihood-free inference and often outperforms competing approaches on standard SBI benchmarks. As part of our contribution, we derive a new Monte Carlo estimator for the posterior distribution when using our DNRE approach. We also derive a simple likelihood gradient estimator that can be successfully used to perform HMC. We are therefore the first to compare random walk MH with HMC for likelihood ratio estimation approahces. We find that HMC is a viable MCMC approach and can outperform random-walk MH. Finally, we introduce a novel application of SBI for the design of a quadcopter and we release this data as part of the supplementary materials.

%% file: appendix.tex
\newpage
\section{Illustrative Gaussian Example: Additional Results}\label{app:ill}

\begin{table}[h!]
\begin{center}
\begin{small}
\begin{sc}
\begin{tabular}{lccc}
\toprule
Approach & MSE & MSE & MSE \\
& $\sigma = 0.1$ & $\sigma = 0.3$ & $\sigma =0.5$ \\
\midrule
NRE & 0.136 & 0.207 & 0.759 \\ 
BNRE & 3.584 & 4.289 & 3.693 \\
DNRE & 0.104 & 0.122 & 0.124 \\\bottomrule
\end{tabular}
\end{sc}
\end{small}
\end{center}
\caption{Displays the mean squared error (MSE) between the ground truth log likelihood ratio and each neural ratio estimator using the results from Figure \ref{fig:gauss}.}
\label{tab:gauss_res}
\end{table}

\section{Summary of Benchmark Description}\label{app:bench}

The benchmark results of Section \ref{sec:bench} are the same as those from the Simulation-Based Inference Benchmark \cite{lueckmann2021benchmarking}. Here we summarize each task by providing the dimensionality and priors but refer to the original benchmark for further details:
\begin{itemize}
    \item \textbf{Two Moons} (\textsc{TM}): $\bm{\theta} \in \mathbb{R}^2$; $\mathbf{x} \in \mathbb{R}^2$; $\bm{\theta} \sim \mathcal{U}(\mathbf{-1}, \mathbf{1})$
    \item \textbf{Gaussian Linear} (\textsc{GL}): $\bm{\theta} \in \mathbb{R}^{10}$; $\mathbf{x} \in \mathbb{R}^{10}$; $\bm{\theta} \sim \mathcal{N}(\mathbf{0}, 0.1\mathbf{I})$
    \item \textbf{Lotka-Volterra} (\textsc{LV}): $\bm{\theta} \in \mathbb{R}^{4}$; $\mathbf{x} \in \mathbb{R}^{20}$; $\theta_1 \sim \text{LogNormal}(-0.125, 0.5)$, $\theta_2 \sim \text{LogNormal}(-3, 0.5)$ $\theta_3 \sim \text{LogNormal}(-0.125, 0.5)$, $\theta_4 \sim \text{LogNormal}(-3, 0.5)$
    \item \textbf{SIR Epidemiological Model} (\textsc{SIR}): $\bm{\theta} \in \mathbb{R}^{2}$; $\mathbf{x} \in \mathbb{R}^{10}$; $\theta_1 \sim \text{LogNormal}(\log (0.4), 0.5)$, $\theta_2 \sim \text{LogNormal}(\log (1/8), 0.2)$
    \item \textbf{Simple Likelihood Complex Posterior} (\textsc{SLCP}): $\bm{\theta} \in \mathbb{R}^{5}$; $\mathbf{x} \in \mathbb{R}^{8}$; $\bm{\theta} \sim \mathcal{U}(\mathbf{-3}, \mathbf{3})$
    \item \textbf{Gaussian Linear Uniform} (\textsc{GLU}): $\bm{\theta} \in \mathbb{R}^{10}$; $\mathbf{x} \in \mathbb{R}^{10}$; $\bm{\theta} \sim \mathcal{U}(\mathbf{-1}, \mathbf{1})$
    \item \textbf{Simple Likelihood Complex Posterior with Distractors} (\textsc{SLCP D}): $\bm{\theta} \in \mathbb{R}^{5}$; $\mathbf{x} \in \mathbb{R}^{100}$; $\bm{\theta} \sim \mathcal{U}(\mathbf{-3}, \mathbf{3})$
    \item \textbf{Bernoulli Generalized Linear Model} (\textsc{B GLM}): $\bm{\theta} \in \mathbb{R}^{10}$; $\mathbf{x} \in \mathbb{R}^{10}$; $\theta_1 \sim \mathcal{N}(0, 2)$, $\bm{\theta}_{2:10} \sim \mathcal{N}(\mathbf{0}, (\mathbf{F}^{\top}\mathbf{F})^{-1})$, $\mathbf{F}_{i,i-2} = 1$, $\mathbf{F}_{i,i-1} = 1$, $\mathbf{F}_{i,i} = 1 + \sqrt{(i-1)/9}$, $\mathbf{F}_{i,j} = 0\ \text{otherwise}$
     \item \textbf{Gaussian Mixture} (\textsc{GM}): $\bm{\theta} \in \mathbb{R}^2$; $\mathbf{x} \in \mathbb{R}^2$; $\bm{\theta} \sim \mathcal{U}(\mathbf{-10}, \mathbf{10})$
     \item \textbf{Bernoulli Generalized Linear Model Raw} (\textsc{B GLM R}): $\bm{\theta} \in \mathbb{R}^{10}$; $\mathbf{x} \in \mathbb{R}^{100}$; $\theta_1 \sim \mathcal{N}(0, 2)$, $\bm{\theta}_{2:10} \sim \mathcal{N}(\mathbf{0}, (\mathbf{F}^{\top}\mathbf{F})^{-1})$, $\mathbf{F}_{i,i-2} = 1$, $\mathbf{F}_{i,i-1} = 1$, $\mathbf{F}_{i,i} = 1 + \sqrt{(i-1)/9}$, $\mathbf{F}_{i,j} = 0\ \text{otherwise}$
\end{itemize}

The \texttt{C2ST} score is a classifier-based test that requires access to ground truth samples from the posterior for each observation, $\mathbf{x}$. In the simulation-based inference benchmark, there are ten observations with corresponding reference posterior samples that allow us to use the \texttt{C2ST} score. We directly use the code from the benchmark which trains a two-layer neural network with a number of units that is ten times the dimension of $\bm{\theta}$. The parameters are normalized and the accuracy is reported using 5-fold cross validation. An accuracy score of 0.5 means that the samples from likelihood-free MCMC are indistinguishable from the ground truth and corresponds to the best performance. 

\subsection{Estimated Posterior under Ground Truth $\bm{\theta}$}\label{app:post}

Table \ref{tab:sbi_log_post} displays the average log posterior probabilities across all observations for each estimator when applied to the ground truth observation and parameter pairs. In addition to seeing that DNRE tends to assign the highest average log posterior probability, we also note that the results provide a further confirmation that our Monte Carlo estimate for DNRE's posterior is working well. For these results we use $10{,}000$ Monte Carlo samples.

\begin{table*}[h!]
\begin{center}
\begin{scriptsize}
\begin{sc}
\begin{tabular}{lccccc}
\toprule
Approach & TM & GL & LV & SIR & SLCP \\
\midrule
NRE & $3.856 \pm 0.447$ & $-0.735 \pm 1.965$ & $9.653 \pm 2.866$ & $7.037 \pm 0.982$ & $-3.99 \pm 1.981$\\
BNRE & $3.778 \pm 0.626$ & $-1.014 \pm 1.966$ & $8.193 \pm 1.949$ & $\mathbf{4.912 \pm 1.133}$ & $-3.995 \pm 1.337$\\
DNRE & $\mathbf{3.701 \pm 0.909}$ & $\mathbf{-0.317 \pm 2.527}$ & $\mathbf{8.035 \pm 1.916}$ & $6.546 \pm 0.685$ & $\mathbf{-2.914 \pm 1.418}$\\
\toprule
& GLU & SLCP D & B GLM & GM & B GLM R \\
\midrule
NRE & $-0.555 \pm 1.395$ & $-9.193 \pm 0.353$ & $-4.392 \pm 2.192$ & $\mathbf{-2.06 \pm 0.542}$ & $-5.109 \pm 2.182$ \\
BNRE & $-0.424 \pm 1.279$ & $-9.303 \pm 0.389$ & $-4.669 \pm 2.534$ & $-2.188 \pm 0.478$ & $-5.124 \pm 1.875$ \\
DNRE & $\mathbf{1.042 \pm 2.829}$ & $\mathbf{-7.54 \pm 1.13}$ & $\mathbf{-3.37 \pm 3.191}$ & $-2.101 \pm 0.519$ & $\mathbf{-4.173 \pm 3.05}$ \\
\bottomrule
\end{tabular}
\end{sc}
\end{scriptsize}
\end{center}
\caption{Log posterior of ground truth parameters, $\bm{\theta}^*$, averaged over reference observations using the SBI Benchmark. The log posterior of DNRE is estimated according to Equation \eqref{eq:dnre_post} with $10{,}000$ Monte Carlo samples.}
\label{tab:sbi_log_post}
\end{table*}

\section{Quadcopter Design}\label{app:uav}
\paragraph{Observations and Parameters} The observation, $\mathbf{x} \in \mathbb{R}^7$ contains: the number of interferences; the mass (Kg); the maximum flight distance (meters); the maximum hover time (seconds); the maximum lateral speed (m/s); the maximum control input at the maximum flight distance ($u \in [0,1]$; and the maximum power at the maximum speed (watts). The quadcoper design parameters, $\bm{\theta} \in \mathbb{R}^{19}$, and priors are given by:
\begin{itemize}
    \item Arm Length (mm) $\sim \mathcal{U}(50, 500)$.
    \item Fuselage Floor Height (mm) $\sim \mathcal{U}(5, 80)$.
    \item Fuselage Horizontal Diameter (mm) $\sim \mathcal{U}(250, 1000)$.
    \item Fuselage Vertical Diameter (mm) $\sim \mathcal{U}(40, 400)$.
    \item Fuselage Autopilot X Offset (mm) $\sim \mathcal{U}(-100, 100)$.
    \item Fuselage Autopilot Y Offset (mm) $\sim \mathcal{U}(-50, 50)$.
    \item Fuselage Battery X Offset (mm) $\sim \mathcal{U}(-100, 100)$.
    \item Fuselage Battery Y Offset (mm) $\sim \mathcal{U}(-50, 50)$.
    \item Fuselage Current X Offset (mm) $\sim \mathcal{U}(-100, 100)$.
    \item Fuselage Current Y Offset (mm) $\sim \mathcal{U}(-50, 50)$.
    \item Fuselage GPS X Offset (mm) $\sim \mathcal{U}(-100, 100)$.
    \item Fuselage GPS Y Offset (mm) $\sim \mathcal{U}(-50, 50)$.
    \item Fuselage Length (mm) $\sim \mathcal{U}(50, 200)$.
    \item Fuselage RPM X Offset (mm) $\sim \mathcal{U}(-100, 100)$.
    \item Fuselage RPM Y Offset (mm) $\sim \mathcal{U}(-50, 50)$.
    \item Fuselage Variometer X Offset (mm) $\sim \mathcal{U}(-100, 100)$.
    \item Fuselage Variometer Y Offset (mm) $\sim \mathcal{U}(-50, 50)$.
    \item Fuselage Voltage X Offset (mm) $\sim \mathcal{U}(-100, 100)$.
    \item Fuselage Voltage Y Offset (mm) $\sim \mathcal{U}(-50, 50)$.
\end{itemize}

\begin{figure}[h!]
    \centering
    \includegraphics[width=\columnwidth]{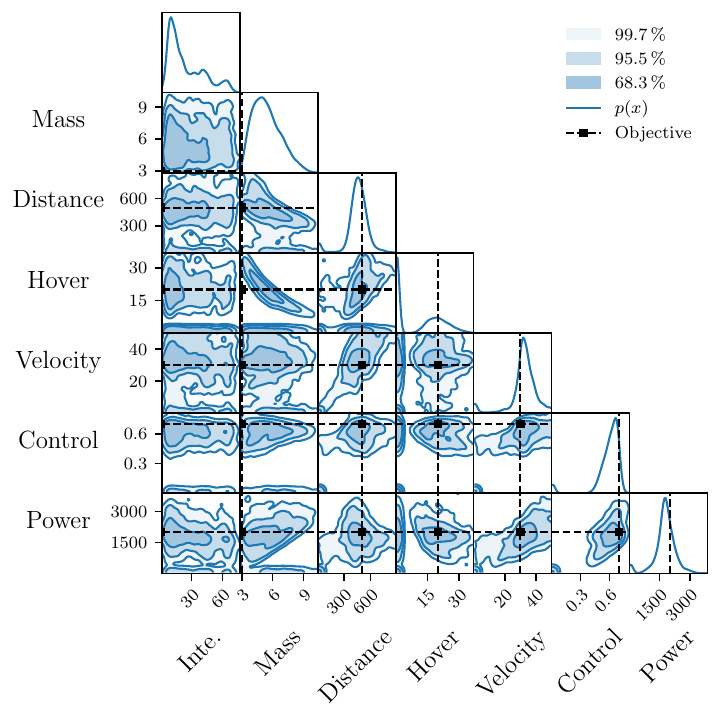}
    \caption{Corner plot displaying the empirical distribution of the observations, $\mathbf{x}$, of the training data as described in Section \ref{sec:uav}. The objective, highlighted by the black square, corresponds to the design specification used in Section \ref{sec:uav}'s experimentation. A notable challenge with this specification can be seen in the Hover vs. Mass density map. The jointly desired mass and maximum hover time falls outside the density of the training data.}
    \label{fig:uav_data}
\end{figure}

\paragraph{Data} Figure \ref{fig:uav_data} displays the empirical distribution of the $4{,}490$ training and validation observations, $\mathbf{x}$. The design objective as specified in Section \ref{sec:uav} is superimposed on the figure. 

\begin{figure}[h!]
    \centering
    \includegraphics[width=0.8\columnwidth]{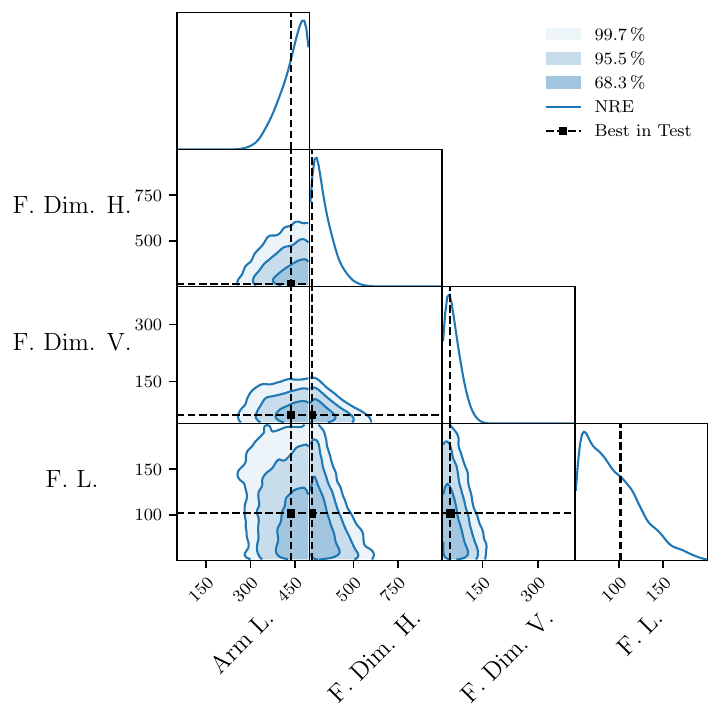}
    \caption{Corner plot displaying materialized samples from likelihood-free inference using NRE for a subset of the parameters as highlighted in Figure \ref{fig:uav}. The black square corresponds to the highest ranked test design in the test set.}
    \label{fig:uav_nre}
\end{figure}

\begin{figure}[h!]
    \centering
    \includegraphics[width=0.8\columnwidth]{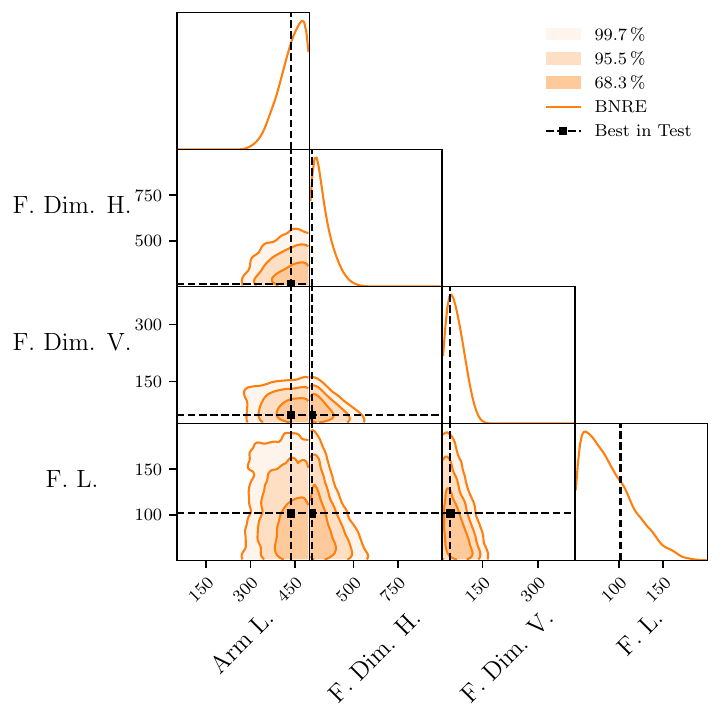}
    \caption{Corner plot displaying materialized samples from likelihood-free inference using BNRE for a subset of the parameters as highlighted in Figure \ref{fig:uav}. The black square corresponds to the highest ranked test design in the test set.}
    \label{fig:uav_bnre}
\end{figure}

\paragraph{Additional results} Complementary to Figure \ref{fig:uav_dnre} in Section \ref{sec:uav}, Figures \ref{fig:uav_nre} and \ref{fig:uav_bnre} show corner plots for likelihood-free MCMC using NRE and BNRE respectively. The distributions are conistent in their shape and the relationships that they display. These figures are plotted using the \textit{LAMPE} simulation-based inference library \cite{lampe}.

\begin{figure}[h!]
    \centering
    \includegraphics[width=\columnwidth]{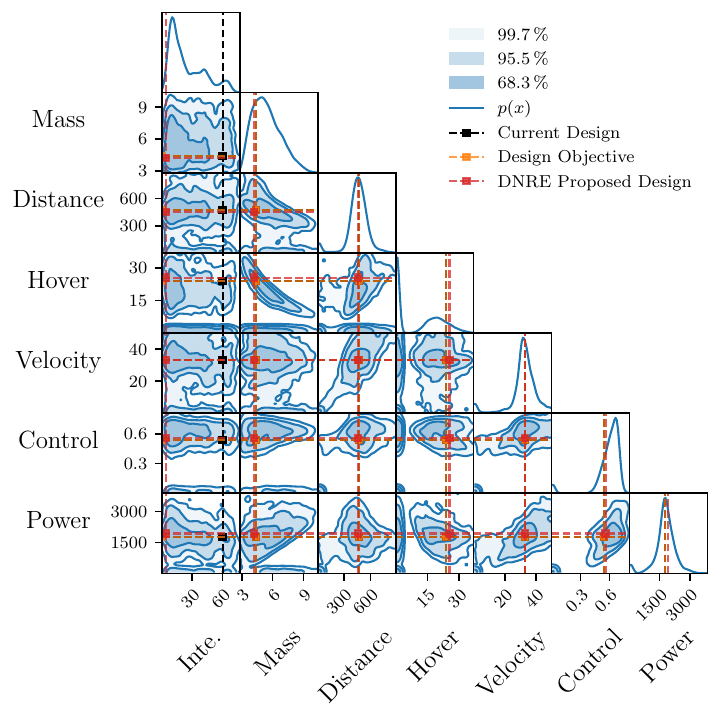}
    \caption{Experiment to improve an existing design. The current design, denoted via the black square, has multiple structural interferences as seen in the first column. Our objective, highlighted in orange, is to keep all the design observations that same while reducing the number of intereferences. The red square is the final sample of an HMC chain using DNRE for performing likelihood-free inference. This last sample is evaluated on the flight simulator and is shown to match closely to the design objective. These points are superimposed on the empirical marginal distribution of the observations for context.}
    \label{fig:uav_dnre_improve}
\end{figure}

\paragraph{Case Study of Improving an Existing Design}
Another interesting use-case of SBI is to use our likelihood ratio estimator to improve an existing design. For example, we may like the performance of an existing quadcopter, but dislike the number of structural interferences. Figure \ref{fig:uav_dnre_improve} shows designs superimposed on the same empirical training distribution as in Figure \ref{fig:uav_data}. Here, the black square is the current design that we are interested in improving. Specifically, the current design has $60$ structural interferences as highlighted in the first column. The other attributes, such as a hover time of $24$~s and a maximum velocity of $33$~m/s, we would like to keep as we reduce the number of interferences.

\begin{figure*}[h!]
    \centering
    \includegraphics[width=\textwidth]{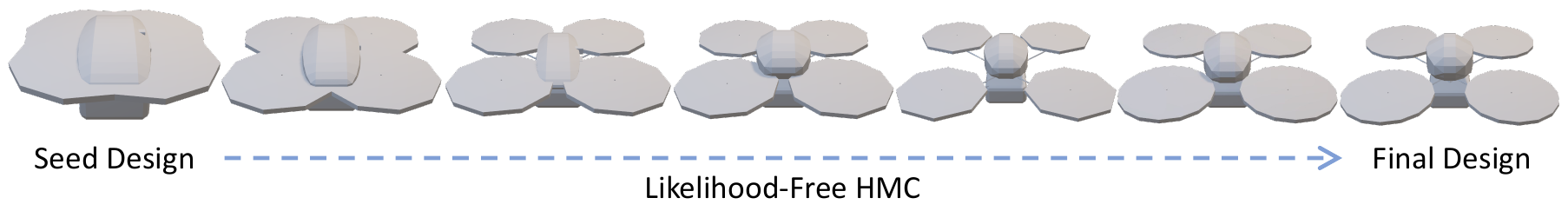}
    \caption{Sub-sampled quadcopter designs taken along the likelihood-free HMC chain using DNRE. The initial seed design on the far left has multiple structural interferences, including sensors that cut through the fuselage. As we move along the chain we see the design morph into our desired structure with very few interferences. This is achieved though increasing the arm length and changing the shape of the fuselage jointly with varying the placement of the interior sensing components.}
    \label{fig:uav_stls_large}
\end{figure*}

As an experiment, we initialize HMC with this initial design and perform likelihood-free inference with DNRE, with a small step size of $0.1$ and a simple trajectory length of $1$. We take $200$ steps with a thinning of $8$ and run each design through the simulator. The final proposed design is shown by the red square in Figure \ref{fig:uav_dnre_improve} and only has $4$ interferences. This closely matches our design objective and further highlights the utility of utilizing HMC for likelihood-free inference, as well as our new direct neural likelihood ratio estimator. Figure \ref{fig:uav_stls_large} directly shows sub-sampled designs that feature along the Markov chain after running the simulator. We see the evolution of our seed design from having a significant number of interferences to almost none. 

\section{Computing Infrastructure}

To reproduce the results of this paper, it is preferable to train the neural estimators using GPUs. In this paper we use an NVIDIA RTX A6000 to train all our models but smaller GPUs can be used since the neural network architectures do not take up significant memory.

